# Learning Image Representations for Content Based Image Retrieval of Radiotherapy Treatment Plans

Charles Huang [1], Varun Vasudevan [3], Oscar Pastor-Serrano [2,4], Md Tauhidul Islam [2], Yusuke Nomura [2], Piotr Dubrowski [2], Jen-Yeu Wang [2], Joseph B. Schulz [2], Yong Yang [2] and Lei Xing [2]

[1] Department of Bioengineering, Stanford University, Stanford, USA
[2] Department of Radiation Oncology, Stanford University, Stanford, USA
[3] Institute for Computational & Mathematical Engineering, Stanford University, Stanford, USA
[4] Department of Radiation Science and Technology, Delft University of Technology, the Netherlands

E-mail: xxx@xxx.xx

**Abstract**

*Objective:* Knowledge based planning (KBP) typically involves training an end-to-end deep learning model to predict dose distributions. However, training end-to-end methods may be associated with practical limitations due to the limited size of medical datasets that are often used. To address these limitations, we propose a content based image retrieval (CBIR) method for retrieving dose distributions of previously planned patients based on anatomical similarity. *Approach:* Our proposed CBIR method trains a representation model that produces latent space embeddings of a patient's anatomical information. The latent space embeddings of new patients are then compared against those of previous patients in a database for image retrieval of dose distributions. All source code for this project is available on github. *Main Results:* The retrieval performance of various CBIR methods is evaluated on a dataset consisting of both publicly available plans and clinical plans from our institution. This study compares various encoding methods, ranging from simple autoencoders to more recent Siamese networks like SimSiam, and the best performance was observed for the multitask Siamese network. *Significance:* Applying CBIR to inform subsequent treatment planning potentially addresses many limitations associated with end-to-end KBP. Our current results demonstrate that excellent image retrieval performance can be obtained through slight changes to previously developed Siamese networks. We hope to integrate CBIR into automated planning workflow in future works, potentially through methods like the MetaPlanner framework.

## 1. Introduction

*1.1 Background*

The workflow for radiotherapy treatment planning typically involves an iterative, trial-and-error process for manually navigating trade-offs (Sethi 2018, Xing et al 1999). Treatment planning optimization contains multiple objectives, which are often conflicting. For this reason, no single plan can optimize performance on all objectives at once, and treatment planning can instead be conceptualized as navigating the set of Pareto optimal, nondominated solutions (Craft et al 2006, 2012, Huang et al 2021).

In an effort to reduce the amount of active planning time in treatment planning, there has been growing interest in automated methods. Many of these methods (such as the MetaPlanner (MP) framework, the Expedited Constrained Hierarchical Optimization (ECHO) system, iCycle, etc.) can be interpreted as navigating the Pareto front while guided by some utility function (Huang *et al* 2022, Zarepisheh *et al* 2019, Breedveld *et al* 2012, Hussein *et al* 2018). While these methods can obtain excellent results, even on complex clinical cases, a few potential limitations remain regarding the design of their utility functions.

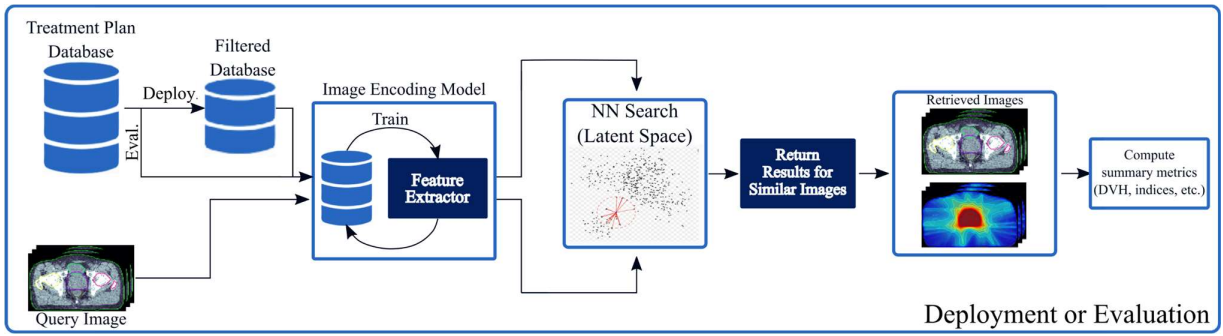

**Figure 1.** Visualizes the workflow for CBIR. Given a new patient during treatment planning (i.e. query image) the method searches a filtered database to retrieve similar images. The corresponding dose distribution can then be used in subsequent automated planning.

Many utility functions used in treatment planning can be broadly categorized as either hand-crafted or data-driven. Hand-crafted utility functions are typically adapted from standard clinical protocols, as well as institutional protocols. In contrast, data-driven utility functions are fundamental to knowledge based planning (KBP), and they typically involve training an end-to-end deep learning model to predict dose distributions or dose volume histograms (DVHs) (Hussein *et al* 2018, Ma *et al* 2019, Babier *et al* 2021, Momin *et al* 2021, Shen *et al* 2020). In principle, using data-driven utility functions has the potential to more closely mimic the decision-making process of human planners. However, practically, there remain many limitations with current end-to-end KBP methods that need to be addressed.

### 1.2 End-to-End Knowledge Based Planning

End-to-end KBP refers to a workflow in automated planning that uses machine learning models to predict dose information (i.e. DVHs, dose distributions, etc.) of new patients without the assistance of other methods. These models have been gaining traction recently in the treatment planning community, in part due to the popularity of machine learning methods in other engineering and computer science fields. However, as with any tool, end-to-end KBP methods have limitations that should warrant further consideration before deployment in clinical settings.

Here, we highlight a few of the main limitations associated with current end-to-end KBP methods. First, end-to-end KBP models are often trained on limited datasets, and their performance may not generalize well between various institutions and protocols. As curating treatment planning data can be prohibitively expensive and public data remains relatively scarce (Babier *et al* 2021), it may not be wise to rely on an end-to-end model entirely. Moreover, dose predictions from models following one protocol typically cannot be used for institutions that use different protocols. Consider, as an example, the Open-KBP dataset, which contains head and neck cases with planning target volumes (PTVs) prescribed doses of 70, 63, and 56 Gy. End-to-end models trained on this dataset cannot be easily applied for the purpose of planning cases using alternative protocols (e.g. using prescription doses of 70, 56, and 52 Gy) (National Cancer Institute (NCI) 2022).

Second, end-to-end KBP methods have no guarantees on the deliverability of their predicted dose distributions. These methods typically perform a pixel-wise least squares regression of the dose distribution, and there are no inherent constraints on the predicted values. For this reason, the predicted dose distributions can be any or all of the following: infeasible, inefficient (not Pareto optimal), or non-compliant with clinical protocols.

### 1.3 Content Based Image Retrieval

In an attempt to address the limitations associated with end-to-end methods, this work proposes a content based image retrieval (CBIR) method that retrieves relevant treatment plans of patients from a database given a new patient's anatomical information (i.e. medical images, contours, etc.), called the query image. CBIR uses a machine learning model to create latent space representations of the query image and images from the database (Latif *et al* 2019, Zin *et al* 2018, Dubey 2021). After computing a distance function (i.e. Euclidean distance) between the query image representation and representations of the database images, we can then sort by the closest distances (Nearest neighbour search) and return the $k$ closest images to the query (Top-$k$ images).

As machine learning methods are heavily influenced by training data, end-to-end methods may yield unusable results when applied to out of distribution cases. This often occurs when training end-to-end models on plans that follow a specific protocol and deploying those models on plans that follow separate protocols (i.e. differing number of PTV levels, differing PTV prescriptions, etc.). Unlike end-to-end methods,

CBIR only utilizes deep learning for image representations. Adapting CBIR to new protocols or guidelines simply involves filtering the database that the retrieval method selects from to only include patients that follow those new desired protocols. Figure 1 provides a visualization of database filtering when deploying CBIR in a clinical setting. For the evaluation or benchmarking purposes of this manuscript, all results will be provided for an unfiltered database. Similarly, while end-to-end methods have no guarantees on the deliverability of their predicted dose distributions, CBIR retrieves plans for previous patients. These previous patient plans were verified to be deliverable and can be filtered to only include those compliant with clinical protocols.

## 2. Methods

### 2.1 Content Based Image Retrieval

Content-based image retrieval (CBIR) aims to search a database for images of similar content (i.e. anatomical information) to a query image. Figure 1 provides the overall CBIR workflow as applied to treatment planning. A database of previous treatment plans is first created and stored. This database contains each patient's anatomical information, which includes their computed tomography (CT) images and relevant contours, as well as their dose distribution. After training the image encoding model, the CBIR method is supplied a new patient's anatomical information, the query image, which it encodes into a latent space embedding that is compared to embeddings of other patients in the database. Image embeddings (i.e. one-dimensional vector representations of each image in the latent space) with the closest Euclidean distance are then retrieved from the database (Nearest neighbour search), and the corresponding dose distribution can be used in subsequent automated planning. During deployment or real-world usage, the database is first filtered to contain plans with the relevant institution and clinical protocols. During all evaluations (i.e. benchmarking for this paper), the unfiltered database is used.

### 2.2 Image Encoding Model

The main task of the image encoding model is to extract features from the provided images. Given images $X \in \mathbb{R}^D$, the goal is to learn an encoding function $f: \mathbb{R}^D \to \mathbb{R}^M$ that produces a continuous latent space embedding $\mathbf{z} \in \mathbb{R}^M$. Here, $X$ refers to a multichannel volume which consists of the CT and contours for each case. For the methods that utilize contrastive learning (i.e. SimSiam and the multitask Siamese network), which are presented in later sections, the input during training also includes a channel for the dose distribution. During deployment and evaluation, the input to all models only includes the CT and contours. The embedding $\mathbf{z}$ refers to the one-dimensional vector representation of images in the latent space. In this work we evaluate the image retrieval performance of five main categories of methods. Readers looking for model design inspiration may find previous reviews of alternative image retrieval tasks to be useful (Zin et al 2018, Dubey 2021, Latif et al 2019).

Prior to training the image encoding model, standard data pre-processing is applied to each patient's images. First, each patient's CT volume, segmentation mask, and corresponding dose distribution are resampled to the dimensions $d = 128 \times 128 \times 128$. The segmentation masks follow a label encoding scheme (with labels ranging from 1 to 4), containing the various planning target volumes (PTVs) and relevant organs-at-risk (OARs). After resampling, each CT volume is then clipped to a soft-tissue window (400 HU width and 0 HU level) and normalized. To keep consistent with common convention in Siamese networks, we interchangeably refer to the input image as the anchor image.

This current work evaluates five main categories of image encoding models used for CBIR: (1) a vanilla autoencoder (Goodfellow *et al* 2016), (2) a variational autoencoder (VAE) (Zhao *et al* 2018), (3) a Siamese network with the triplet margin loss (Schroff *et al* 2015), (4) SimSiam (Chen and He 2020), and (5) a multitask Siamese network (Caruana 1997, Schroff *et al* 2015, Chen and He 2020). For the encoder portion of all evaluated models, we utilize the same backbone convolutional neural network (CNN) architecture (consisting of multiple convolution, GroupNorm, and LeakyReLU blocks). Similarly, the latent space embedding vector has a size that is empirically set to 1024 for all models.

### 2.2.1 Vanilla Autoencoder

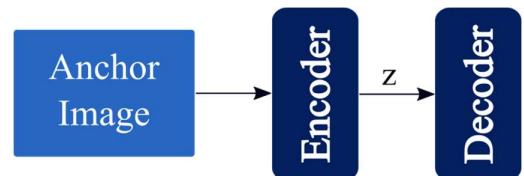

**Figure 2a.** A schematic of the vanilla autoencoder model architecture.

The vanilla autoencoder consists of a standard CNN encoder and a transposed convolution decoder (Goodfellow *et al* 2016). Figure 2a provides a schematic of the model.

### 2.2.2 Information Maximizing Variational Autoencoder

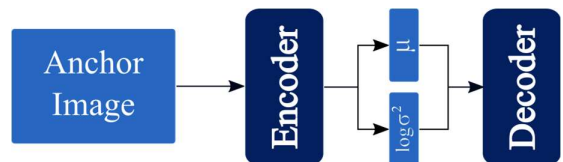

**Figure 2b.** A schematic of the Information-maximizing Variational Autoencoder model architecture. Outputs of the encoder are vectors describing the mean and variance of the latent space distributions.

The information maximizing variational autoencoder (Info-VAE) model is a generative model which uses an additional maximum mean discrepancy (Gretton *et al* 2006) objective, as proposed by Zhao et al. (Zhao *et al* 2018). As the InfoVAE is a generative model, the typical embedding vector $z$ is a random variable sample using the reparameterization trick (Kingma and Welling 2013, Zhao *et al* 2018). In order to get a deterministic output, we follow the common practice of using the "μ" layer output as the latent space embedding vector instead of the vector $z$.

Figure 2b provides a schematic of the InfoVAE architecture, and the loss function for the InfoVAE model is listed in Equation 1.

$$L_{\text{Info-VAE}} = -E_{z \sim q(Z|X)}[log\, p_\theta(x|z)] + (1-\alpha)D_{KL}(q_\theta(z|x)||p(z)) + (\alpha + \lambda - 1)D_{MMD}(q_\theta(z|x)||p(z)) \quad (1)$$

Here, the first term refers to the reconstruction loss (typically mean squared error), the second term $D_{KL}$ refers to the Kullback–Leibler (KL) divergence, and the third term $D_{MMD}$ refers to the maximum mean discrepancy (Gretton *et al* 2006). $p(z)$ refers to the prior, $q_\theta(z|x)$ refers to the variational posterior, $p_\theta(x|z)$ refers to the true posterior, α and λ are hyperparameters controlling the amount of regularization, and θ refers to the parameters of the network (Kingma and Welling 2013, Zhao *et al* 2018).

### 2.2.3 Siamese Network with a Triplet Loss Function

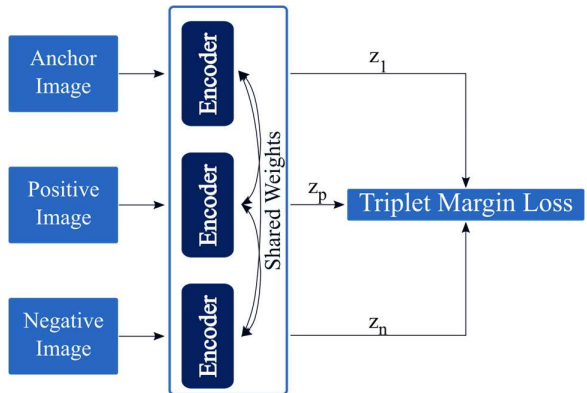

**Figure 2c.** A schematic of the Siamese network with a triplet loss function.

The Siamese network with a triplet loss (SNTL) is a classic method used for CBIR (Chechik *et al* 2010, Schroff *et al* 2015). The Siamese network refers to a network which contains duplicate encoders, where each shares parameters with its duplicates. The triplet loss function is provided in Equation 2. To construct triplets, we take a sample image from the dataset (i.e. anchor image). The positive image can then be sampled by taking another image from the same class (see the Dataset Section) as the anchor image, while the negative

$$L_{Triplet} = max(||z_1 - z_p||_2 - ||z_1 - z_n||_2 + margin, 0) \quad (2)$$

image refers to an image of a different class than the anchor image. Figure 2c provides a schematic of the SNTL model architecture. The triplet loss function computes a distance between the embeddings for the anchor image $z_1$ and positive image $z_p$, as well as a distance between the embeddings for the anchor image and negative image $z_n$. The goal is then to make the distance between the anchor and positive embeddings at least some margin smaller than the distance between the anchor and negative embeddings.

### 2.2.4 Simple Siamese Network

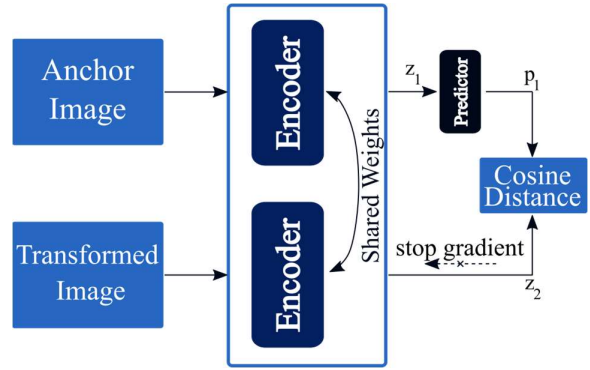

**Figure 2d.** A schematic of the SimSiam network, which uses a stop gradient and the cosine distance. Here the transformed image refers to the dose distribution of patients during model training. Only the anchor image is required for deployment and evaluation.

The simple Siamese (SimSiam) network (Chen and He 2020) is a recent representation learning method that extends on previous state of the art methods like SimCLR (Chen *et al* 2020a) and BYOL (Grill *et al* 2020). SimSiam uses a stop-gradient to learn meaningful representations without the use of negative sample pairs, large batches, or momentum encoders. As these are usually difficult to obtain, utilizing an approach like SimSiam can be more practical than other recent representation learning methods. Figure 2d provides a schematic of the SimSiam model, and the loss function is listed in Equation 3. In order to incorporate information from the dose distributions during model training, we set the transformed image as a multichannel input that uses the dose distribution and contour information. This training scheme is also used by the multitask approach described below.

$$L_{SimSiam} = -\frac{p_1}{||p_1||_2} \cdot \frac{z_2}{||z_2||_2} \quad (3)$$

Here, $p_1 = h(f(x_1))$ refers to the output of the predictor $(h)$, $z_1$ refers to the embedding of the anchor image, and $z_2$ refers to the embedding of the transformed image.

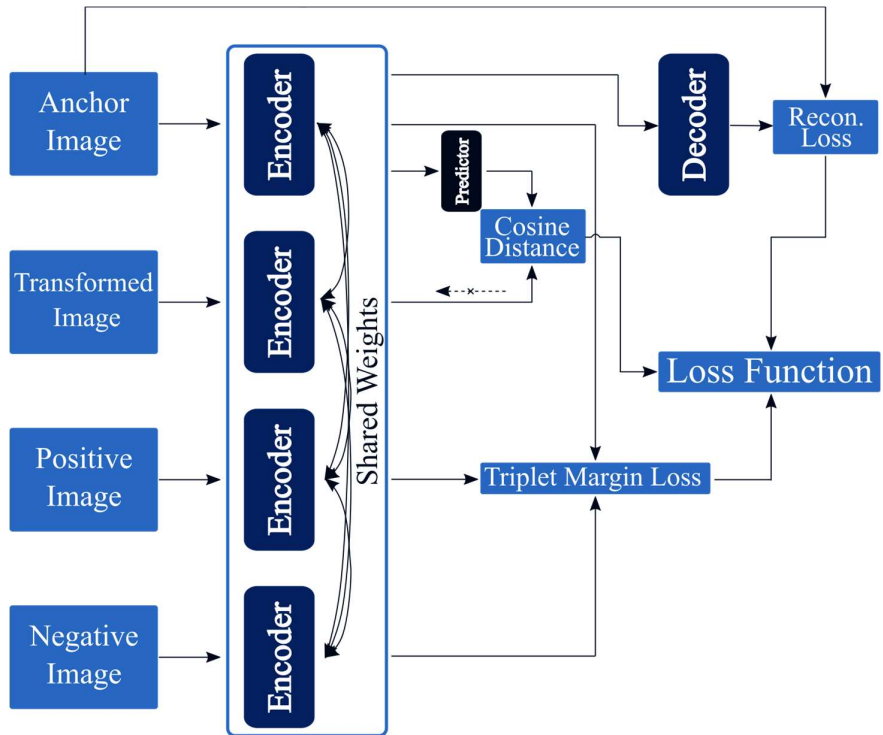

**Figure 2e.** A schematic of the multitask Siamese network, which combines tasks from many of the previously discussed approaches.

## 2.2.5 Multitask Siamese Network

Following the typical multitask learning scheme (Caruana 1997), the multitask Siamese network (MSN) combines many of the previously mentioned approaches. Due to the small dataset size of this study, the MSN attempts to improve generalization by utilizing information from the reconstruction task, SimSiam embedding task, and the triplet loss task. Figure 2e provides a schematic of the MSN model, and the loss function is provided in Equation 4. In this study, $\beta$ and $\gamma$ were empirically set to values of $1e-2$ and $1e-1$, respectively.

$$L_{multitask} = L_{\text{recon}} + \beta L_{\text{SimSiam}} + \gamma L_{Triplet} \tag{4}$$

## 2.3 Dataset

The dataset used in this study contains 405 cases composed of public data (OpenKBP) (Babier *et al* 2021) and private data, which were collected as part of clinical workflow. The body sites included in this dataset are prostate and head and neck, with either volumetric modulated arc therapy (VMAT) or intensity modulated radiation therapy (IMRT) used for treatment.

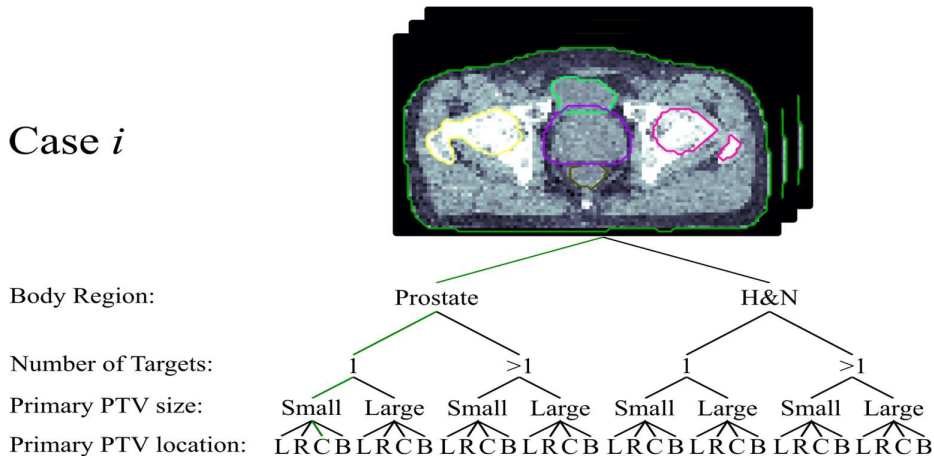

**Figure 3.** Visualizes the workflow for classifying an example patient for benchmark evaluation purposes.

For evaluation, all cases in the dataset were manually classified according to the following four criteria for a total of 32 classes:

1. Which body site does the case belong to (i.e. prostate or head and neck)?
2. How many target levels are there?
3. Is the primary PTV small or large?
4. How is the primary PTV located (i.e. left, right, center, or bilateral)?

Figure 3 provides a visualization of the workflow taken to classify patients in the dataset. Cases were split into 235 in the training phase, 43 in the validation phase, and 127 in the testing phase. All source code for this project has been made available on github (https://github.com/chh105/MetaPlanner/tree/main/cbir).

## 3. Results

### 3.1 Evaluation

Performance of the included image retrieval methods was evaluated for three aspects: retrieval performance, clustering performance, and qualitative performance. We begin by retrieving $k$ images from the database that have embeddings closest to that of the query image. Retrieval performance can then be evaluated using standard metrics like the classification accuracy, precision, recall, and $F_1$-score of the top-$k$ retrieved images (Mogotsi 2010). In this study, we show results for $k$ ranging from 1 to 5, though smaller or larger ranges of $k$ may also be used. The definitions for these evaluation metrics are listed in Table 1. Moreover, clustering performance is then evaluated using standard metrics like the cluster homogeneity, completeness, v-measure, adjusted Rand index, and adjusted mutual information (Rosenberg and Hirschberg 2007, Hubert and Arabie 1985, Steinley 2004, Strehl and Ghosh 2003). Lastly, qualitative performance is evaluated by visually inspecting the retrieval results for example query patients.

**Table 1.** Definitions for various retrieval (multiclass) evaluation metrics. Each metric is computed for the top-$k$ images returned by the retrieval systems.

| | Definition |
|---|---|
| Accuracy@k | $f(k) = \frac{1}{N_{classes}} \sum_{i=1}^{N_{classes}} \frac{tp + tn}{tp + fp + fn + tn}$ |
| Precision@k | $f(k) = \frac{1}{N_{classes}} \sum_{i=1}^{N_{classes}} \frac{tp}{tp + fp}$ |
| Recall@k | $f(k) = \frac{1}{N_{classes}} \sum_{i=1}^{N_{classes}} \frac{tp}{tp + fn}$ |
| $F_1 - score@k$ | $f(k) = \frac{2 \cdot Precision \cdot Recall}{Precision + Recall}$ |
| Retrieval Score | $s = \sum_{k=1}^{n} f(k) \cdot \left(\frac{1}{2}\right)^k$ |

tp := true positives, tn := true negatives
fp := false positives, fn := false negatives

### 3.2 Image Retrieval Performance

We first evaluate the image retrieval performance of the candidate image encoding models using the metrics in Table 1. Figure 4 plots accuracy, precision, recall, and F-score as functions of $k$. Ground-truth labels for each patient are provided by following the procedure described in Section 2.3. For each metric, we can compute a simple retrieval scoring function by applying an exponential weighting to each retrieval metric. Here, greater emphasis is placed on small values of $k$, as only the most relevant retrieved plans would be used to inform subsequent treatment planning. Values of the retrieval scoring functions are listed in Table 2.

For all retrieval scores, MSN achieves the best performance. The second-best performance for all retrieval scores is achieved by the SNTL method, followed by the more recent SimSiam method. Performances for the vanilla autoencoder and Info-VAE were comparable, suggesting that the variational loss component may not be entirely useful for our image retrieval task.

### 3.3 Clustering Performance

**Table 2.** Performance of the candidate encoding models is evaluated in regards to retrieval and clustering. Retrieval scores are computed using an exponential weighting of each metric as a function of $k$. Cluster metrics are computed using standard formulas, where $k$=1.

| | Accuracy Retrieval Score | Precision Retrieval Score | Recall Retrieval Score | $F_1$ Retrieval Score | Homogeneity | Completeness | V-measure | Adjusted Rand Index | Adjusted Mutual Info. |
|---|---|---|---|---|---|---|---|---|---|
| Multitask Siamese Network | **1.23** | **1.04** | **1.03** | **1.02** | **0.683** | **0.679** | **0.681** | **0.516** | **0.593** |
| Info VAE | 0.70 | 0.58 | 0.57 | 0.52 | 0.438 | 0.427 | 0.432 | 0.275 | 0.278 |
| Siamese Network (Triplet Loss) | 1.11 | 0.96 | 0.97 | 0.95 | 0.671 | 0.653 | 0.662 | 0.437 | 0.571 |
| SimSiam (Cosine Similarity) | 0.78 | 0.80 | 0.59 | 0.60 | 0.412 | 0.422 | 0.417 | 0.247 | 0.265 |
| Vanilla Autoencoder | 0.72 | 0.62 | 0.56 | 0.56 | 0.372 | 0.375 | 0.374 | 0.194 | 0.204 |

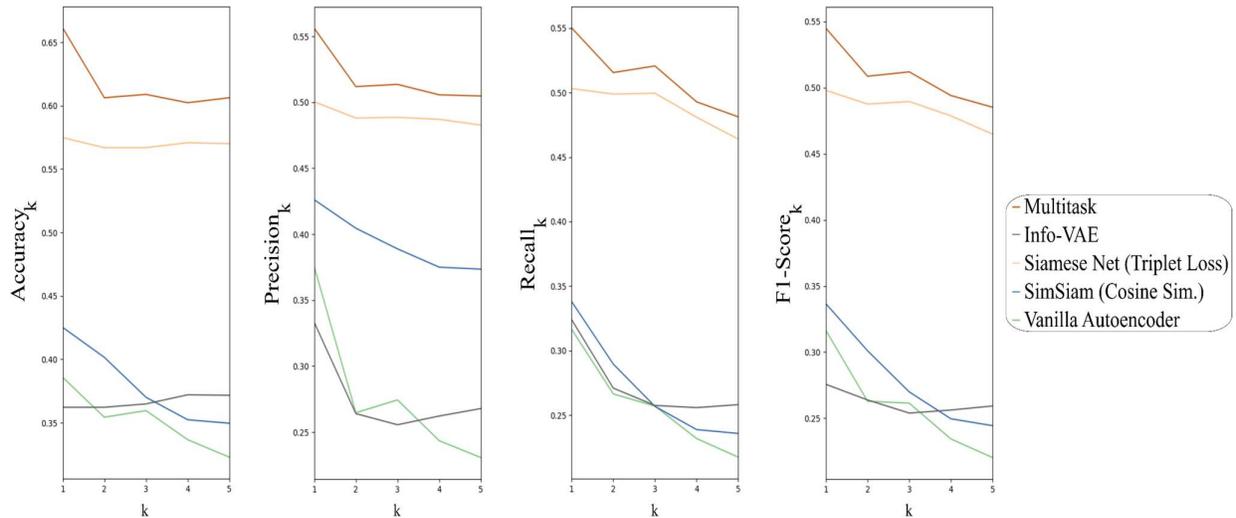

**Figure 4.** Plot of the retrieval metrics for the top-k images. For all retrieval metrics, best performance was achieved using the multitask Siamese network.

We additionally evaluate clustering performance using metrics listed in the last 5 columns of Table 2 (Rosenberg and Hirschberg 2007, Hubert and Arabie 1985, Steinley 2004, Strehl and Ghosh 2003). Each score is computed on the latent space embeddings produced by the candidate methods, where ground truth labels are provided following the procedure detailed in Section 2.3 and prediction labels are computed for $k = 1$.

Cluster homogeneity (Rosenberg and Hirschberg 2007) assesses the ability to create clusters that contain only members of a single class. Cluster homogeneity is bounded between [0,1], and performance of an ideal method approaches a cluster homogeneity of 1. Cluster completeness (Rosenberg and Hirschberg 2007) assesses the ability to assign all members of a class to the same cluster. Cluster completeness is bounded between [0,1], and performance of an ideal method approaches a cluster completeness of 1. V-measure (Rosenberg and Hirschberg 2007) is computed as the harmonic mean of cluster homogeneity and completeness. V-measure is bounded between [0,1], and performance of an ideal method also approaches a value of 1. The adjusted Rand index (Hubert and Arabie 1985) measures the similarity between ground truth class assignments and those of the clustering method, adjusted for chance groupings. The adjusted Rand index is bounded between $[-1,1]$, and performance of an ideal method approaches a value of 1. Finally, the adjusted mutual information (Strehl and Ghosh 2003) measures the agreement between ground truth class assignments and those of the clustering method, adjusted for chance groupings. The adjusted mutual information is bounded between $[0,1]$, and performance of an ideal method approaches a value of 1.

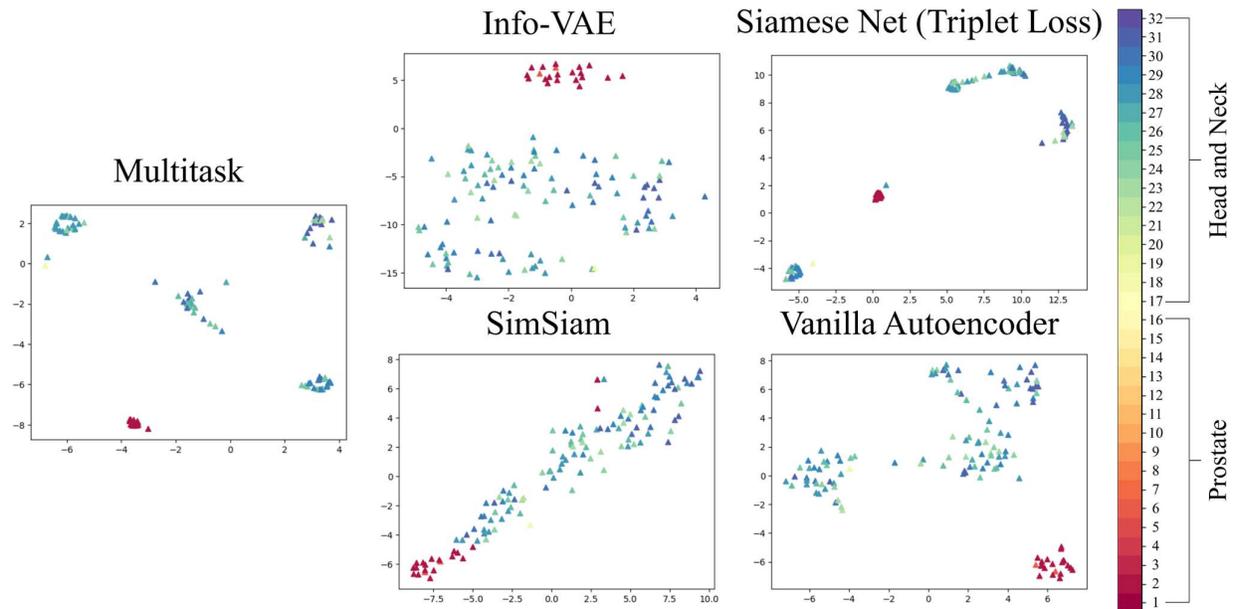

**Figure 5.** Visualization of the latent space embeddings of the query images computed by each of the candidate encoding models.

Of the benchmarked encoders, the top three performers for cluster homogeneity, cluster completeness, V-measure, adjusted Rand index, and adjusted mutual information were the MSN, SNTL, and Info VAE models (Table 2). Figure 5 shows a TSNE plot of the latent space embeddings for the query set, with the evaluation class labels provided for visualization purposes (ground truth labels for all evaluations are computed following Section 2.3). Here, embeddings for the MSN are substantially more distinct and grouped than those of the other candidate models.

*3.4 Qualitative Performance*

A qualitative comparison of retrieved images for an example query image is provided in Figure S1 of the Supplemental Materials. This example query image is a head and neck case, has multiple PTV levels, and has a large primary PTV located on the right-side of the patient. Both the MSN and SNTL models retrieve patients of the same classification from the database. Moreover, the retrieved patient for the MSN model is more anatomically similar to the query than that of the SNTL model. The remaining models did not retrieve patients from the database of the same classification as the query image. For all evaluations in this current work, retrieval is performed on the unfiltered database. However, during practical deployment, the database will first be filtered by the relevant classification.

**4. Discussion**

Here, a CBIR framework is used to retrieve treatment plans from a database. The proposed CBIR framework compares the latent space embeddings of a query image to those of images in a database for the purpose of image retrieval. To produce latent space embeddings, we evaluate various encoding models in regards to retrieval performance, clustering performance, and visual quality.

End-to-end KBP methods rely solely on a machine learning model to create dose predictions for new patients. As these end-to-end methods are often trained using limited datasets, they may become unusable for cases that come from different institutions and protocols, due to poor generalizability. Instead of relying entirely on learning-based approaches, it may be more practical to utilize machine learning as a tool only for certain tasks in the dose prediction process, but not for the entire pipeline. In contrast to end-to-end methods, CBIR only uses an encoding model to produce latent space embeddings of each patient's anatomical information. Consequently, errors accrued from poor model performance have less of an impact on the predicted dose distributions than those accrued when performing an end-to-end approach. During CBIR deployment, we can additionally filter the database to contain only relevant plans from specified institutions or protocols, thereby ensuring that the retrieved treatment plans are protocol compliant and deliverable. Given the current limitations with data availability in treatment planning, approaches like CBIR can be deployed as more practical, robust alternatives to end-to-end methods.

The proposed CBIR framework retrieves relevant treatment plans from a database and can be utilized in any pipeline that would otherwise incorporate end-to-end KBP. Specifically, the retrieved dose distributions can be used in methods which directly optimize machine parameters through dose mimicking (Eriksson and Zhang 2022, McIntosh *et al* 2017, Mahmood *et al* 2018). They can be alternatively used in modular methods like the MetaPlanner framework (Huang *et al* 2022), which optimize treatment planning hyperparameters and can be more robust than direct dose mimicking.

In this work, various candidate encoder models were evaluated to determine viable CBIR options. Of the evaluated methods, the multitask Siamese network consistently performed the best in regards to retrieval performance, clustering performance, and visual quality. The dataset used in this study includes a total of 405 cases. Though this may be considered sizeable in the context of medical data, it certainly cannot compare to datasets used routinely in computer vision (Deng *et al* 2009, Lecun *et al* 1998). Given the relatively small dataset size used in the current study, the multitask model manages to outperform its alternatives by incorporating additional loss function terms to reduce overfitting. This is evident when observing the performance of methods like SNTL, SimSiam, or the vanilla autoencoder, which individually do not perform as well as the multitask model.

In future work, we plan to incorporate the proposed image retrieval method into automated planning workflow. To clarify the various components for such a process, we describe an example implementation using image retrieval to create a data-driven utility function for automated planning (Figure 6). In this example, a CBIR system is deployed to retrieve a reference plan (i.e. dose distribution, DVHs, etc.) from the database based on similarity to the query patient. We can then compute a distance metric (utility function) between the reference plan and the query patient's current plan in order to guide automated planning. Currently, many automated planning methods model the treatment planning process as one where planners navigate the trade-offs of Pareto optimal solutions using a hand-crafted utility function. However, incorporating the retrieved dose distributions from an image retrieval approach enables the use of data-driven utility functions for automated planning, potentially providing a better model of the decision-making process.

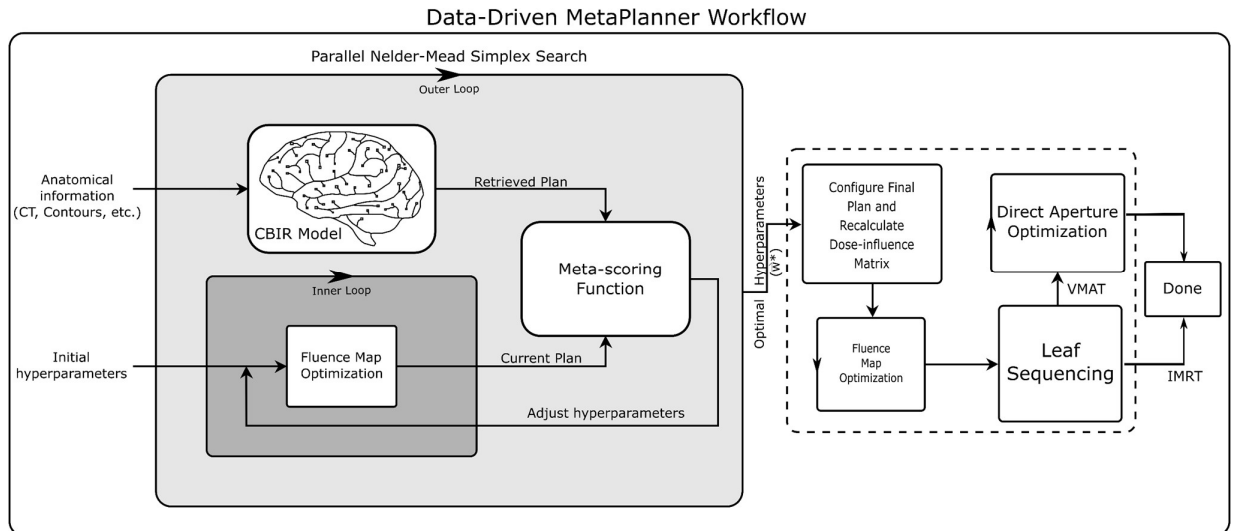

**Figure 6.** Visualization of the workflow for deploying CBIR in automated planning using the MetaPlanner framework.

This study is additionally subject to some limitations. First, while several candidate methods for encoding images were evaluated here, there may certainly exist better performing encoding models that were not tested. Second, due to data availability, we were not able to evaluate other body sites such as lung data, liver data, etc.

External beam radiation therapy is a highly popular treatment modality (Bilimoria *et al* 2008). Recently, there has been growing interest in developing automated methods for the radiotherapy pipeline. Deep learning has generally been successful in performing radiotherapy tasks like segmentation, outcome prediction, etc. (Boldrini *et al* 2019, Liang *et al* 2021, Yuan *et al* 2019, Chen *et al* 2020b, Nomura *et al* 2020, Dong and Xing 2020, Pastor-Serrano and Perkó 2021, 2022), and applying learning based methods to treatment planning also has potential. In future works, we hope to apply the proposed CBIR method directly to automated planning, potentially through frameworks like MetaPlanner (Huang *et al* 2022). Similarly, we plan to address some of the mentioned limitations of the current study by evaluating alternative CBIR encoding models and utilizing data from additional body sites.

## 5. Conclusion

In this work, we introduced a CBIR method to inform subsequent treatment planning. The proposed workflow addresses some key limitations present in traditional end-to-end KBP methods, including generalizability, deliverability, and protocol compliance of predicted dose distributions. To determine a viable encoding model for CBIR, we evaluated several methods ranging from the Info-VAE to Siamese networks with various loss functions to a multitask network that combines tasks from other candidate approaches. Our results indicate that the multitask encoding model consistently provides the best performance when evaluated in regards to retrieval performance, clustering performance, and visual quality.

## Acknowledgements

This research was supported by the National Institutes of Health (NIH) under Grants 1R01 CA176553, R01CA227713, and T32EB009653, Varian Medical Systems, as well as a faculty research award from Google Inc.

**Supplemental Materials**

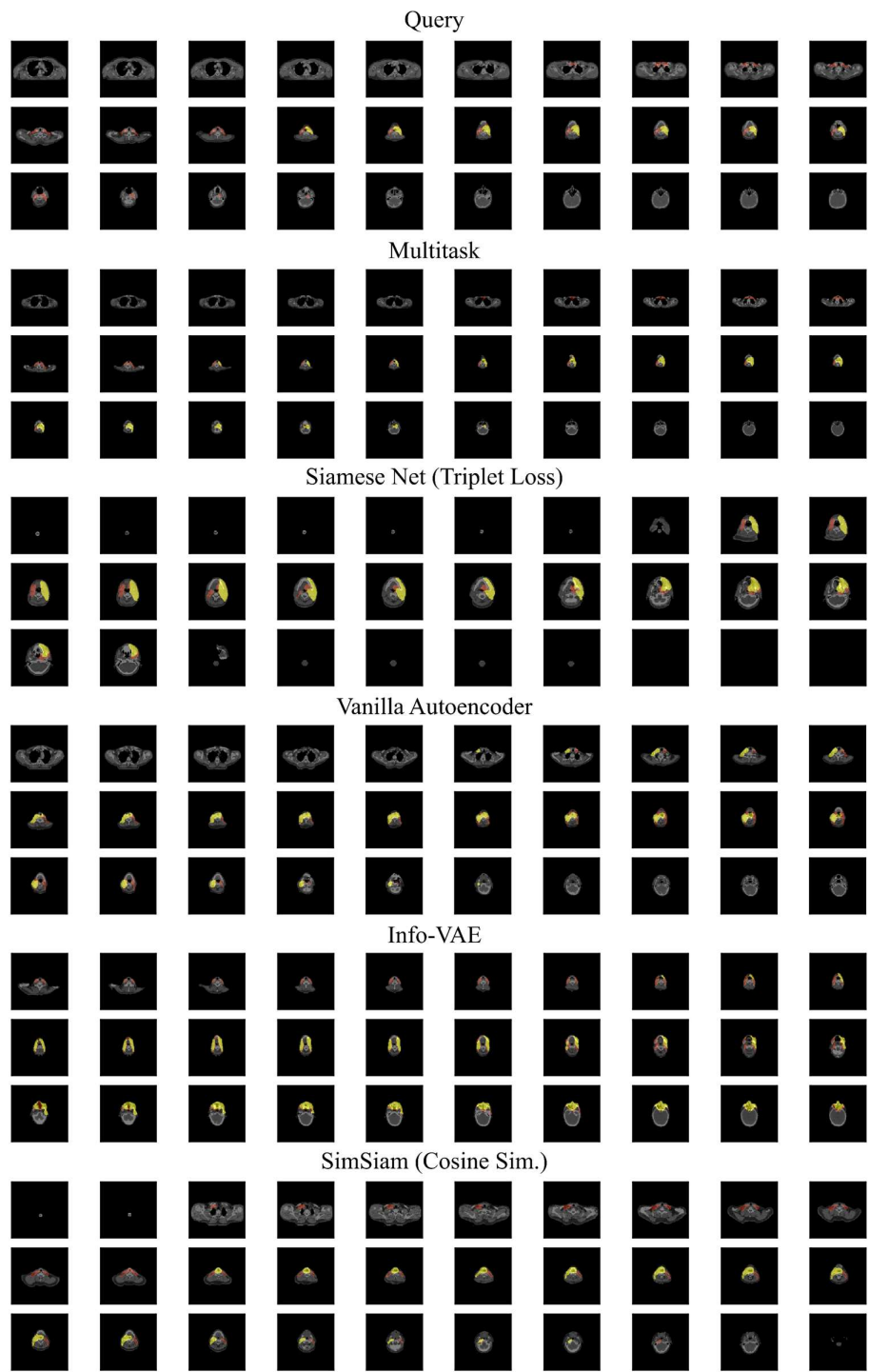

**Figure S1.** Provides a visualization of the retrieval results for an example query image, which belongs to a class of head and neck cases where there are multiple target levels and the primary PTV is located on the right side of the patient. The multitask Siamese network retrieves a patient of the same class as the query image that is also anatomically similar. The SNTL model also retrieves a patient of the same class, though the retrieved image has less anatomical similarity to the query (CT volume does not start in the thoracic region). For this query example, the remaining methods do not retrieve an image of the same class.